%% file: paper.tex
\documentclass{article}
\usepackage{spconf,amsfonts,amsmath,amssymb,graphicx}
\usepackage[T1]{fontenc}
\usepackage[acronym]{glossaries}
\usepackage{multirow}
\usepackage{paralist}
\usepackage{booktabs}
\usepackage{url}
\usepackage{setspace}
\usepackage{siunitx}

\usepackage{array}
\newcommand{\PreserveBackslash}[1]{\let\temp=\\#1\let\\=\temp}
\newcolumntype{C}[1]{>{\PreserveBackslash\centering}p{#1}}
\newcolumntype{R}[1]{>{\PreserveBackslash\raggedleft}p{#1}}
\newcolumntype{L}[1]{>{\PreserveBackslash\raggedright}p{#1}}

\input{bibliography.tex}

\title{Leveraging Multilingual Self-Supervised Pretrained Models for Sequence-to-Sequence End-to-End Spoken Language Understanding}
\name{Pavel Denisov, Ngoc Thang Vu}
\address{Institute for Natural Language Processing (IMS), University of Stuttgart, Germany}
\copyrightnotice{979-8-3503-0689-7/23/\$31.00~\copyright2023 IEEE}
\begin{document}
%\ninept
%
\newacronym{slu}{SLU}{Spoken Language Understanding}
\newacronym{asr}{ASR}{Automatic Speech Recognition}
\newacronym{nlu}{NLU}{Natural Language Understanding}
\newacronym{e2e}{E2E}{End-to-End}
\newacronym{ssl}{SSL}{Self-Supervised Learning}
\newacronym{ic}{IC}{Intent Classification}
\newacronym{sf}{SF}{Slot Filling}
\newacronym{ner}{NER}{Named Entity Recognition}
\newacronym{aed}{AED}{Attention Encoder-Decoder}
\newacronym{ctc}{CTC}{Connectionist Temporal Classification}
\newacronym{mc}{MC}{Modality Correlation}
\newacronym{cer}{CER}{Concept Error Rate}
\newacronym{cver}{CVER}{Concept/Value Error Rate}

\maketitle
\begin{abstract}
A number of methods have been proposed for End-to-End Spoken Language Understanding (E2E-SLU) using pretrained models, however their evaluation often lacks multilingual setup and tasks that require prediction of lexical fillers, such as slot filling. In this work, we propose a unified method that integrates multilingual pretrained speech and text models and performs E2E-SLU on six datasets in four languages in a generative manner, including the prediction of lexical fillers. We investigate how the proposed method can be improved by pretraining on widely available speech recognition data using several training objectives. Pretraining on 7000 hours of multilingual data allows us to outperform the state-of-the-art ultimately on two SLU datasets and partly on two more SLU datasets. Finally, we examine the cross-lingual capabilities of the proposed model and improve on the best known result on the PortMEDIA-Language dataset by almost half, achieving a Concept/Value Error Rate of 23.65\%.
\end{abstract}

\vspace{-5pt}
\section{Introduction}
\gls{slu} is a common name for tasks combining speech and language processing to extract semantic concepts from spoken sentences, such as intents, slots, and named entities. This functionality is essential to various systems with a voice interface, including intelligent assistants and automatic call answering services.
Traditionally, \gls{slu} has been decomposed to \gls{asr} and \gls{nlu} subtasks that are solved
sequentially in a pipeline manner. In this scenario, \gls{asr} converts speech recording
to text representation that is then processed by \gls{nlu}. The advantage of
a pipelined approach is that both \gls{asr} and \gls{nlu} can be optimized independently
using numerous datasets labelled for the corresponding tasks. The disadvantages are that:
first, the text representation lacks paralinguistic information, such as prosody
and punctuation in most cases, and in addition to that
contains errors introduced by \gls{asr} and propagated to \gls{nlu};
and second, sequential execution of \gls{asr} and \gls{nlu}
introduces a time lag that is not desirable in an interactive context.
These downsides motivated the community to work on
the \gls{e2e} \gls{slu} methods that allow building a single model performing
\gls{slu} task directly on speech input.

One of the critical challenges in \gls{e2e} \gls{slu} is data sparsity since
the availability of labelled datasets for \gls{slu} is even lower
than for \gls{asr} or \gls{nlu}. Transfer learning is a popular technique
that alleviates the data sparsity problem for one task by learning
from another related task. More recently, \gls{ssl} methods
demonstrate the possibility of gaining improvements even without requiring annotation
for any task by training on unlabelled data of corresponding modalities,
such as text \cite{lewis2020bart} or speech \cite{baevski2020wav2vec}.

Usually, \gls{slu} tasks are treated as a classification problem,
either on the sequence level (\gls{ic} \cite{denisov2020pretrained}) or on the token level
(\gls{sf} \cite{bastianelli2020slurp} and \gls{ner} \cite{shon2022slue}).
Few works, however, represent \gls{slu} tasks as generation
problems \cite{rongali2021exploring,pelloin2021end2end}.
There are multiple examples of classification-based \gls{slu} systems
utilizing \gls{ssl} pretrained speech and text encoders \cite{denisov2020pretrained,laperriere2023use}.
However, usage of \gls{ssl} pretrained models in generation-based \gls{slu} systems
is limited to speech encoders \cite{arora2022espnet}.
Aside from that, generation-based \gls{slu} descriptions are typically focused
on one language only \cite{he2023interpreter,sunder2023fine}.
We aim to close these two gaps by employing multilingual \gls{ssl} pretrained
speech and text models for solving \gls{slu} tasks in multiple languages
via sequence-to-sequence modeling.
In this work, we propose a unified architecture for building \gls{e2e} \gls{slu}
models and evaluate it on a diverse set of established \gls{slu} benchmarks.
Our experiments demonstrate that multilingual \gls{ssl} pretrained text-to-text model
can be fine-tuned to solve token level \gls{nlu} tasks in a generative way and this can be transferred to speech
modality with the help of multilingual \gls{ssl} pretrained speech encoder.
Furthermore, we improve the \gls{nlu} to \gls{slu} transferability by aligning the
hidden representations of speech and text using a medium-sized multilingual
\gls{asr} dataset and different training approaches: \gls{ctc}, \gls{aed} and
a novel \gls{mc} objective. In several cases, our results are better
than the best previously reported.
We provide the implementation, configurations, data preparation and
scoring scripts and pretrained models at \url{https://github.com/DigitalPhonetics/multilingual-seq2seq-slu}.

\begin{figure*}[!t]
\centering
\includegraphics[scale=1.0]{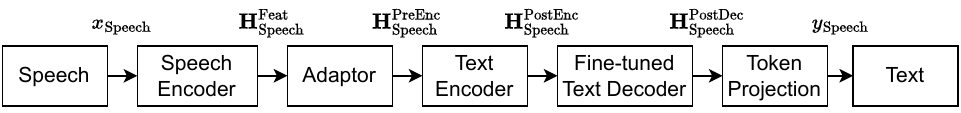}
\caption{
	General architecture of our SLU model.
	The speech encoder and text encoder are initialized from unchanged SSL pretrained models.
	The text decoder is initialized from an SSL pretrained model
	that is fine-tuned on the ground truth transcriptions of an SLU dataset.
	The Adaptor is trained from scratch on an SLU dataset or is pretrained
	on ASR data (section \ref{sec:pretraining}).
	}
\label{fig:model}
\end{figure*}

\section{Method}

\subsection{SLU model}
\label{sec:model}
The proposed approach is outlined in Figure \ref{fig:model}.
Our \gls{slu} model combines \gls{ssl} pretrained speech encoder
and text-to-text encoder-decoder models.
The raw speech input $x$ is processed by the speech encoder that outputs
an acoustic representation $\textbf{H}^{\mathrm{Feat}}$.
The Adaptor maps the acoustic representation $\textbf{H}^{\mathrm{Feat}}$
from the speech encoder to a quasi-graphemic representation $\textbf{H}^{\mathrm{PreEnc}}$
that resembles text token embeddings and is fed to the text encoder.
Output of the text encoder $\textbf{H}^{\mathrm{PostEnc}}$
is decoded by the autoregressive text decoder producing an output
hidden representation $\textbf{H}^{\mathrm{PostDec}}$ that is
projected to text token logits used to produce an output text token sequence $y$.
The speech and text encoder layers are initialized from the parameters of
the general pretrained models without any changes.
The text decoder parameters are first fine-tuned on the corresponding \gls{slu} dataset
using the ground truth transcriptions as input and the \gls{slu} annotations as output,
thus resulting in the text based \gls{nlu} model.
The text encoder is kept frozen during this step to ensure the transferability
to the \gls{slu} model initialized from the general parameters.
Only the Adaptor's parameters have to be learned during the \gls{slu} training,
and this motivates us to investigate the Adaptor pretraining.

\vspace{-10pt}
\begin{figure*}[htb]
	\begin{minipage}[b]{.50\linewidth}
		\centering
		\centerline{\includegraphics[scale=0.71]{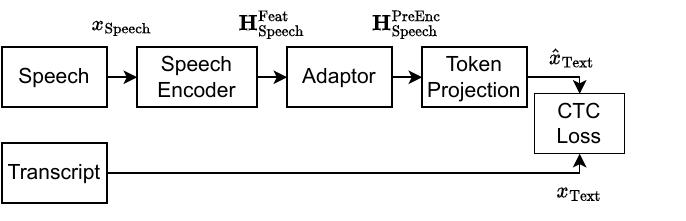}}
		%  \vspace{1.5cm}
		\centerline{(a) PreEnc CTC Adaptor pretraining.}\medskip
	\end{minipage}
	\hfill
	\begin{minipage}[b]{0.48\linewidth}
		\centering
		\centerline{\includegraphics[scale=0.71]{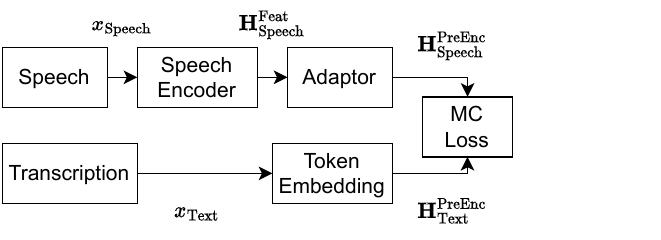}}
		%  \vspace{1.5cm}
		\centerline{(b) PreEnc MC Adaptor pretraining.}\medskip
	\end{minipage}
	\begin{minipage}[b]{.46\linewidth}
		\centering
		\centerline{\includegraphics[scale=0.71]{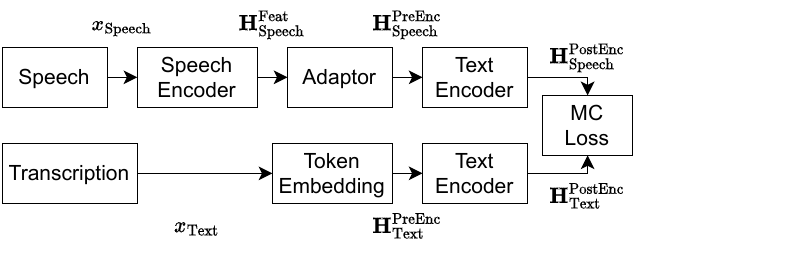}}
		%  \vspace{1.5cm}
		\centerline{(c) PostEnc MC Adaptor pretraining.}\medskip
	\end{minipage}
	\hfill
	\begin{minipage}[b]{.52\linewidth}
		\centering
		\centerline{\includegraphics[scale=.71]{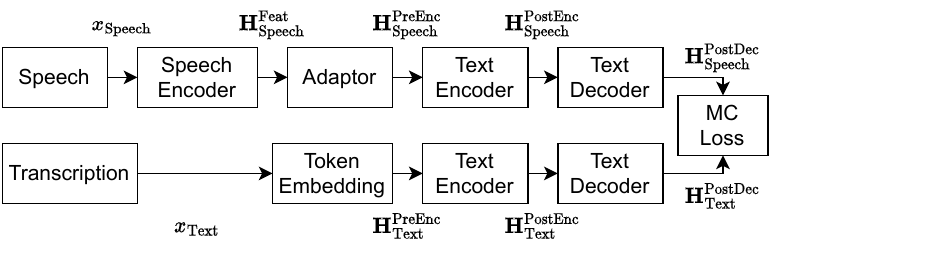}}
		%  \vspace{1.5cm}
		\centerline{(d) PostDec MC Adaptor pretraining.}\medskip
	\end{minipage}
	\begin{minipage}[b]{0.98\linewidth}
		\centering
		\centerline{\includegraphics[scale=.71]{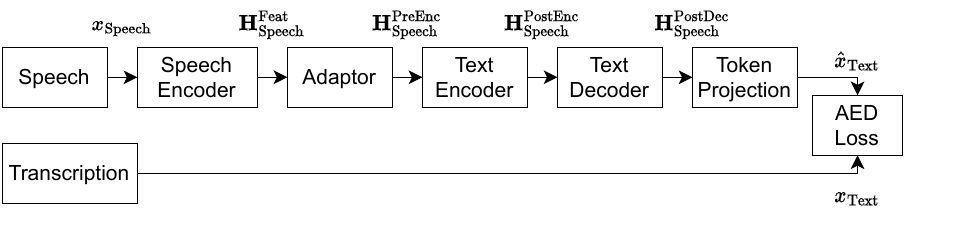}}
		%  \vspace{1.5cm}
		\centerline{(e) PostEnc AED Adaptor pretraining.}\medskip
	\end{minipage}
\vspace{-10pt}
	\caption{
		Investigated options for the Adaptor pretraining using pairs of
		speech recordings and text transcriptions.
		The Adaptor is optimized to predict the hidden representation of the speech that is
		as close to the hidden representation of the text as possible.
		Each Adaptor pretraining option refers to a combination
		of the hidden representation layer (PreEnc, PostEnc, PostDec)
		and the training method (MC, CTC, AED).
		The MC loss can be applied directly to the hidden representations of the speech and
		the text before the text encoder (b), after the text encoder (c) or after the text decoder (d).
		Alternatively, the hidden speech representation can be projected to the token logits
		via a regular linear transformation (Token Projection)
		allowing CTC training if the hidden speech representation is extracted before the text encoder (a)
		or AED training if the hidden speech representation is extracted after the text decoder (e).
	}
	\label{fig:pretraining}
\end{figure*}

\subsection{Adaptor pretraining}
\label{sec:pretraining}
Generally, the Adaptor pretraining aims to minimize the distance between
the speech representation in the \gls{slu} model and the text representation
in the \gls{ssl} text model. As Figure~\ref{fig:pretraining} shows, this can
be done on multiple levels of the neural network using various training strategies.
We select three types of a text representation and the corresponding levels
of the \gls{slu} model to extract a logically similar speech representation:
\begin{inparaenum}[(i)]
    \item text token embeddings $\textbf{H}^{\mathrm{PreEnc}}_{\mathrm{Text}}$
    and the \gls{slu} Adaptor output $\textbf{H}^{\mathrm{PreEnc}}_{\mathrm{Speech}}$;
    \item the hidden text representation $\textbf{H}^{\mathrm{PostEnc}}_{\mathrm{Text}}$ after the encoder of the \gls{ssl} text model
    and the hidden speech representation $\textbf{H}^{\mathrm{PostEnc}}_{\mathrm{Speech}}$ after the text encoder of the \gls{slu} model;
    \item the output text representation $\textbf{H}^{\mathrm{PostDec}}_{\mathrm{Text}}$ after the decoder of the \gls{ssl} text model
    and the hidden speech representation $\textbf{H}^{\mathrm{PostDec}}_{\mathrm{Speech}}$ after the text decoder of the \gls{slu} model.
\end{inparaenum}

The Adaptor output can be aligned with the text token embeddings using the \gls{ctc} loss function
in the vein of LegoNN approach \cite{dalmia2022legonn}. While this is the most straightforward
and computationally cheap way to pretrain Adaptor parameters in our framework,
it might set an unnecessary strict target for the alignment because
the text-to-text network has some level of robustness to noisy inputs
and not all differences between text and speech representations
are equally harmful for the prediction of the correct output.

Given this consideration, it might be better to pretrain the Adaptor in the full
\gls{aed} model while keeping all parameters except of the Adaptor in a frozen state.
This way, the output of the decoder serves as a source of training signal
and the Adaptor parameters receive only the relevant updates for the prediction of the correct output.
The disadvantage of this approach is that
the differences between the original decoder
and a task specific \gls{nlu} decoder used in a \gls{slu} model
can hinder the transferability of the pretrained Adaptor output.
It is fair to assume that the practical importance of this problem
depends on the actual differences between the original and
task specific decoders.
These differences can be avoided
by the parameter efficient tuning of the decoder \cite{li2021prefix}.

Finally, any hidden speech representation can be trained directly on a hidden text
representation as a target. In order to do that, we propose the modality matching
approach.
Its design is inspired by the cross-modal grounding methods \cite{zhang2021explainable,khorrami2021evaluation}
and the Barlow twins loss \cite{zbontar2021barlow}.
First, we construct a speech-text correlation matrix:
\begin{equation*}
\textbf{C}_{\mathrm{Speech-Text}} = \textbf{H}_{\mathrm{Speech}} {(\textbf{H}_{\mathrm{Text}})}^\intercal,
\end{equation*}
where $\textbf{H}_{\text{Speech}} \in \mathbb{R}^{L_\mathrm{max} \times d_\mathrm{model}}$ and
$\textbf{H}_{\text{Text}} \in \mathbb{R}^{L_\mathrm{max} \times d_\mathrm{model}}$
are padded normalized hidden speech and text
representations, $L_\mathrm{max} = max(L_\mathrm{Speech}, L_\mathrm{Text})$
is the maximum length of the speech and text sequences and
$d_\mathrm{model}$ is the hidden representation dimension in both modalities.
The representations are zero-padded and normalized to Euclidean norm along
each hidden representation vector.
We mask out the loss for elements outside of $L_\mathrm{Text}$,
because the zero-dominated matrices caused convergence problems.
This can be reduced to the truncation instead of the padding and masking,
but we implement it that way because of the batching.
A ground truth for $\textbf{C}_{\mathrm{Speech-Text}}$ is set to be analogous
text-text self-correlation matrix:
\begin{equation*}
\textbf{C}_{\mathrm{Text-Text}} = \textbf{H}_{\mathrm{Text}} {(\textbf{H}_{\mathrm{Text}})}^\intercal
\end{equation*}
During the training, we minimize the mean square error between
$\textbf{C}_{\mathrm{Speech-Text}}$ and $\textbf{C}_{\mathrm{Text-Text}}$.
The \gls{mc} approach can be applied to the encoder outputs and
therefore allows a trade-off between the two previously mentioned levels of the neural network.
In addition, it can be employed as an alternative to the hard label based
\gls{ctc} and \gls{aed} approaches applicable only at the specific levels of the neural network.

\begin{table}[!t]
\vspace{-10pt}
    \caption{SLU benchmarks used for the evaluation. \#Concepts denotes number of unique slot labels for SF or unique entity labels for NER.}
    \label{tab:datasets_slu}
    \center
    \scriptsize
\setlength{\tabcolsep}{1.8mm}
    \begin{tabular*}{1\columnwidth}{c | c | c | c | c | c | c}
    \toprule
     & SLURP & SLUE & CATSLU & MEDIA & PM-Dom & PM-Lang \\
    \midrule
    \multirow{3}{*}{\parbox{.9cm}{\centering Samples (train/ dev/test)}} & 120K/ & 5.0K/ & 7.0K/ & 12.0K & 5.8K/ & 6.0K/ \\
     & 8.6K/ & 1.7K/ & 1.6K/ & 1.2K/ & 1.3K/ & 1.9K/ \\
     & 13.0K & 1.8K  &  5.8K & 3.5K  & 2.8K & 3.8K \\
    \midrule
    \multirow{3}{*}{\parbox{.9cm}{\centering Hours (train/ dev/test)}} & 84.7/ & 14.0/ & 5.5/ & 16.1/ & 7.4/ & 7.5/ \\
     & 6.9/  & 4.9/ & 1.3/ & 1.6/ & 1.6/ & 2.5/ \\
     & 10.3  & 4.9 & 4.8 & 4.8 & 3.4 & 5.0 \\
    \midrule
    \multirow{1}{*}{\parbox{.9cm}{\centering Language}} & English & English & Mandarin & French & French & Italian \\
    \multirow{1}{*}{\parbox{.9cm}{\centering Tasks}} & IC, SF & NER & SF & SF & SF & SF \\
    \multirow{1}{*}{\parbox{.9cm}{\centering \#Concepts}} & 56 & 7 & 54 & 143 & 34 & 124 \\
    \bottomrule
  \end{tabular*}
\end{table}

\section{Experimental setup}

\subsection{Data}
We assess the performance of the proposed method using the six established SLU benchmarks:
SLURP \cite{bastianelli2020slurp}, SLUE-VoxPopuli \cite{shon2022slue} (evaluating on the validation set),
CATSLU \cite{zhu2019catslu}, MEDIA \cite{devillers2004french},
PortMEDIA-Domain and PortMEDIA-Language \cite{lefevre2012leveraging}.
These datasets cover three SLU tasks in four languages belonging to three families
and have medium to low resource data regimes.
Details of the SLU datasets are given in Table~\ref{tab:datasets_slu}.
In addition to SF annotation, each utterance of SLURP is labeled with one out of 59 unique intents.
We evaluate the \gls{slu} systems using the established metrics for each dataset:
\gls{ic} accuracy and SLU-F1 for SLURP, F1 and label-F1 for SLUE-VoxPopuli,
accuracy and F1 for CATSLU, \gls{cer} and \gls{cver} for MEDIA,
PortMEDIA-Domain and PortMEDIA-Language datasets.
The SLU-F1 combines the slot label F1 with the word and character edit distances
of the slot value \cite{bastianelli2020slurp}.
The \gls{cer} and \gls{cver} are based on the edit distance and are calculated the same way
as the word error rate, but on the sequences of slot labels or slot labels combined with the values instead
of words.

The Adaptor pretraining experiments are performed using subsets
of Common Voice Corpus 9.0 \cite{ardila2020common} and WenetSpeech \cite{zhang2022wenetspeech}.
We select the 36 languages that are present in both XLS-R \cite{babu2021xls}
and mBART50 \cite{tang2020multilingual} pretraining data
and sample down the resulting training and validation subsets uniformly
to 1000 and 20  hours respectively.

\vspace{-5pt}
\subsection{Training details}

\textbf{\gls{nlu} models} are obtained by fine-tuning
the SSL text-to-text model on the ground truth
transcriptions of each \gls{slu} dataset and its task specific outputs.
After preliminary experiments the mBART50 Large model \cite{tang2020multilingual}
was chosen because of its best scores.
According to the original mBART50 approach,
the language is encoded as a special token that
is added at the beginning of the input sequence and
is given as the initial output token to the decoder.
Parameters of the model's encoder and token embeddings are frozen.

\textbf{\gls{slu} models} are implemented in ESPnet-SLU toolkit \cite{arora2022espnet} and follow its
SLURP recipe. We use the weighted-sum of hidden states \cite{yang2021superb,chang2021exploration}
of XLS-R (0.3B) pretrained model \cite{babu2021xls} as speech features. The Adaptor module is
technically organized as a VGG/Conformer based encoder \cite{gulati2020conformer,guo2021recent}
followed by a convolutional Length Adaptor \cite{li2021multilingual}.
This design is based on the encoder architecture
of ASR \cite{chang2021exploration} and SLU \cite{arora2022espnet} systems
and demonstrated better results in our preliminary experiments
compared to the direct fine-tuning of \gls{ssl} speech encoders.
Output of the Adaptor module is fed to the general mBART50 text encoder that is followed
by the fine-tuned text decoder from the \gls{nlu} model.
As in \gls{nlu}, both the text encoder and decoder are conditioned
on the \gls{slu} dataset language by
adding the special language token embedding at the beginning
of the text encoder's input sequence and
by setting the special language token as the initial output of the text decoder.
Conformer layers are configured with $d_\mathrm{model}=1024$,
$d_\mathrm{ff}=4096$, $d_h=8$, $E=8$ and $\mathrm{Conv}$ kernel size of 31.
The Length Adaptor contains a 1-dimensional convolutional layer with stride 2
and reduces the length of input sequence by factor of 2.
Label smoothing with a penalty of 0.1, as well as 3-way speed
perturbation \cite{ko2015audio} data augmentation method are
utilized during the training.
The training is done with Adam optimizer
\cite{adam} with $\beta_1=0.9$, $\beta_2=0.999$, $\epsilon=10^{-8}$
and warmup learning rate scheduler. Number of epochs and warmup steps, maximum learning rate
and batch size are tuned individually for each \gls{slu} dataset.

\textbf{SLU Adaptor pretraining} is carried out
with the configuration similar to the \gls{slu} model training.
We set the maximum learning rate to 5e-5, the number of warmup steps
to 25k and the number of epochs to 30 with the early stopping after 3
epochs of no improvements in validation accuracy.

\begin{table}[]
    \caption{
    	Results of the \gls{nlu} models on the ground truth transcriptions
	and the \gls{slu} models on the speech recordings (without Adaptor pretraining).
    }
    \label{tab:baseline}
    \center
    \footnotesize
    \begin{tabular*}{0.6\columnwidth}{l | l | c  c }
    \toprule
    Dataset & Metrics  & NLU & SLU  \\
    \midrule
    \multirow{2}{*}{SLURP} & IC Acc.$\uparrow$ & 85.67 & 86.97 \\
                           & SLU-F1$\uparrow$ & 79.30  & 77.71 \\
    \midrule
    \multirow{2}{*}{SLUE} & F1$\uparrow$ &  83.25 & 68.90 \\
                          & label-F1$\uparrow$ & 87.76 & 82.28 \\
    \midrule
    \multirow{2}{*}{CATSLU} & Acc.$\uparrow$ & 82.56 & 63.87  \\
                            & F1$\uparrow$ & 73.48  &  48.33 \\
    \midrule
    \multirow{2}{*}{MEDIA} & CER$\downarrow$ &  16.50 & 13.67  \\
                           & CVER$\downarrow$ &  19.09  &  16.28 \\
    \midrule
    \multirow{2}{*}{PM-Dom} & CER$\downarrow$ &  23.49 & 21.43  \\
                            & CVER$\downarrow$ &  26.59  &  24.62 \\
    \midrule
    \multirow{2}{*}{PM-Lang} & CER$\downarrow$ &  40.76  &  25.13 \\
                             & CVER$\downarrow$ &  43.93  &  29.11 \\
    \bottomrule
  \end{tabular*}
\end{table}

\section{Results}

\subsection{Baseline}
Table \ref{tab:baseline} shows the results of the \gls{nlu} models
on the ground truth transcriptions and the baseline \gls{slu}
results on the speech recordings.
Each baseline \gls{slu} system is essentially the corresponding \gls{nlu}
model transferred to the speech modality using the \gls{ssl} pretrained speech
encoder as described in the section \ref{sec:model}.
The Adaptor's parameters in the baseline \gls{slu} systems
have to be trained from scratch using the speech recordings
from the \gls{slu} datasets only.
First of all, we note that the SSL
text-to-text model can in principle be fine-tuned to perform
various \gls{nlu} tasks in a generative manner.
Next, we observe that the gap between the \gls{nlu} and \gls{slu}
performance correlates with the
amount of training data for the English benchmarks, which suggests
that the low resource setting is an issue for our initial \gls{slu} model.
On the other side, \gls{slu} performance is slightly better
than the \gls{nlu} performance for the French benchmarks and
is much better for the Italian benchmark,
despite the small amount of training data. This might be due to the encoder
or token embeddings freeze during the \gls{nlu} training:
both French and Italian likely have smaller presence
in the pretraining data of the original mBART50 model.
Finally, the Mandarin benchmark demonstrates the large gap between
the \gls{nlu} and \gls{slu} performance, what we explain by
the very small presence of Mandarin and related languages
in the XLS-R pretraining.

\subsection{Adaptor pretraininig}

\begin{table*}[]
    \caption{
    	Results depending on the layer and loss function for Adaptor pretraining
	and the average Relative Error Reduction (RER) compared to the error rate without Adaptor pretraining.
	The error rate is calculated by subtracting from 100 for the accuracy, SLU-F1, F1 and label-F1 metrics.
	The last two lines contain the results for the combined losses.
	\underline{Underlined} numbers show the best scores for the individual losses (without combination).
	\textbf{Bold} numbers show the best overall scores.
	}
    \label{tab:results_loss}
    \center
    \footnotesize
    \begin{tabular*}{1.031\textwidth}{l | l | c  c | c  c | c  c | c  c | c  c | c  c | c}
    \toprule
    ID & Adaptor & \multicolumn{2}{c|}{SLURP} & \multicolumn{2}{c|}{SLUE} & \multicolumn{2}{c|}{CATSLU} &
    		   \multicolumn{2}{c|}{MEDIA} & \multicolumn{2}{c|}{PM-Dom} & \multicolumn{2}{c|}{PM-Lang} & Average  \\
    \cmidrule{3-14}
     & pretraining  & \scriptsize \gls{ic} Acc.$\uparrow$ &
    		   \scriptsize SLU-F1$\uparrow$ &
		   \scriptsize F1$\uparrow$ &
		   \scriptsize label-F1$\uparrow$ &
                   \scriptsize Acc.$\uparrow$ &
		   \scriptsize F1$\uparrow$ &
	           \scriptsize CER$\downarrow$ &
		   \scriptsize CVER$\downarrow$ &
		   \scriptsize CER$\downarrow$ &
		   \scriptsize CVER$\downarrow$ &
		   \scriptsize CER$\downarrow$ &
		   \scriptsize CVER$\downarrow$ & RER, \% \\
    \midrule
    0 & None        & 86.97 & 77.71 & 68.90 & 82.28 & 48.33 & 63.87 & 13.67 & 16.28 & 21.43 & 24.62 & 25.13 & 29.11 & - \\
    \midrule
    1 & PreEnc MC  & 87.79 & 77.85 & 69.60 & 83.60 & 43.63 & 59.02 & 13.65 & 16.66 & 20.51 & 24.02 & 26.49 & 30.62 & -3.15 \\
    2 & PreEnc CTC & 88.59 & 78.80 & 71.25 & 84.04 & 50.33 & 65.73 & 12.44 & 15.13 & 19.54 & 23.06 & 22.93 & 26.50 & 6.81 \\
    3 & PostEnc MC & 88.13 & 78.91 & 71.93 & 85.10 & 47.50 & 62.65 & \underline{12.31} & \underline{15.07} & 19.67 & 22.81 & 23.36 & 27.22 & 5.12 \\
    4 & PostDec MC & 88.93 & 79.22 & \underline{72.95} & 86.56 & 50.52 & 65.84 & 12.60 & 15.25 & \underline{\textbf{17.90}} & \underline{\textbf{21.15}} & 22.52 & 26.06 & 9.95 \\
    5 & PostDec AED & \underline{\textbf{89.33}} & \underline{\textbf{80.08}} & 72.94 & \underline{\textbf{86.57}} & \underline{\textbf{54.54}} & \underline{\textbf{69.24}} & 13.64 & 16.18 & 18.00 & 21.27 & \underline{\textbf{21.88}} & \underline{\textbf{25.04}} & \underline{\textbf{12.63}} \\
    \midrule
    6 & 2 + 5 & 89.14 & 80.07 & \textbf{73.07} & 86.05 & 53.63 & 68.37 & 12.12 & \textbf{14.71} & 19.38 & 22.36 & 23.28 & 26.78 & 10.97 \\
    7 & 2 + 3 + 5 & 89.20 & 79.58 & 73.05 & 86.20 & 50.81 & 65.55 & \textbf{12.11}  & 14.83 & 18.85 & 21.91 & 23.40 & 26.72 & 9.17 \\
    \bottomrule
  \end{tabular*}
\end{table*}

\begin{figure}[]
\center
\includegraphics[trim={45pts 32pts 30pts 45pts},clip,width=0.65\columnwidth]{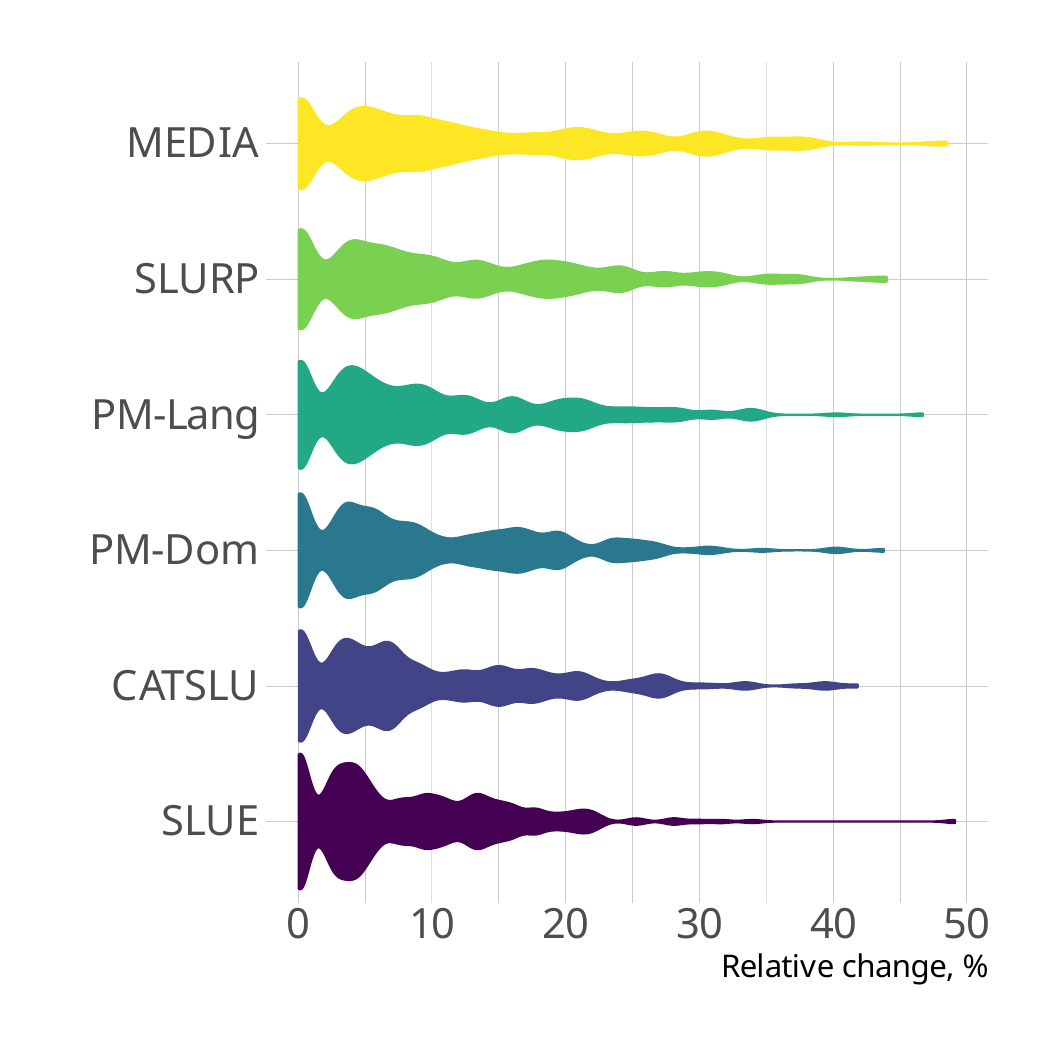}
\caption{
Distribution of the differences between the parameters of the original mBART50
decoder and the fine-tuned NLU decoders depending on the dataset.
The difference is computed as a percentage of the original parameter value.
A wider line indicates a larger number of parameters having certain difference.
}
\label{fig:decoder}
\end{figure}

A comparison of the Adaptor pretraining methods is given in Table \ref{tab:results_loss}.
It is evident from these results that both medium and low resource benchmarks benefit
from the Adaptor pretraining in general.
Hard label loss appears to yield better performance
than the \gls{mc} for all datasets on the PreEnc level and
for three out of six datasets on the PostDec level.
Interestingly, the PostDec level provides the best results on all datasets except MEDIA.
Four out of these datasets contain the smallest amount of training sentences,
therefore the changes in the decoder during the \gls{nlu} fine-tuning
are probably sufficiently small so that it is compatible with the Adaptor
pretrained with the general mBART50 decoder.
SLURP dataset also works best with the PostDec AED pretrained Adaptor despite
the largest amount of training sentences.
Unlike the rest of the datasets, we observe the best scores on MEDIA
for the PostEnc MC aligned Adaptor,
possibly because the \gls{nlu} fine-tuning on this task leads to
the larger modifications in the mBART50 decoder, therefore reducing
its compatibility with the general decoder.
Visualization of the parameter differences in Figure \ref{fig:decoder} shows that
the relative change lies in the higher range for the higher number of parameters
after fine-tuning on MEDIA dataset compared to the other tasks.
Additionally, we calculate the ratio of unique tokens from each SLU dataset that is seen during the pretraining,
as shown in Table \ref{tab:datasets_tokens}. It can be observed from these numbers
that the AED method is more effective for the datasets with at least 95\%
unique test set tokens seen during the pretraining and
the MC loss is more effective otherwise. The MEDIA dataset again stands out in this
analysis with the lowest ratio of unique training set tokens seen in the pretraining,
this might also explain the lower PostDec pretraining effectiveness.

After exploring the training approaches separately,
we try to combine them by simply
summing the loss outputs.
No systems with the multitask pretrained Adaptor degrade considerably compared to
the best system for each dataset, which indicates that different
pretraining objectives regularize each other when combined.
Although the multitask pretraining does not outperform any single objective pretraining,
the combination of CTC and AED can be recommended
as a trade-off solution when little is known about the SLU data.

\begin{table}[]
    \caption{
    	Influence of the SLU benchmark's unique tokens seen in the Adaptor pretraining data
	on the best pretraining method for that SLU benchmark.
	}
    \label{tab:datasets_tokens}
    \center
    \footnotesize
    \begin{tabular*}{0.9\columnwidth}{l | C{0.15\columnwidth} C{0.15\columnwidth} | l}
    \toprule
    Dataset & \multicolumn{2}{c|}{Tokens seen in pretraining, \%} & Best pretraining  \\
    \cmidrule{2-3}
       &  Train & Test & method \\
    \midrule
    SLURP & 98.45 & 98.93 & PostDec AED \\
    SLUE  & 98.93 & 99.66 & PostDec MC/AED \\
    CATSLU & 98.93 & 99.41 & PostDec AED \\
    MEDIA & 91.59 & 93.38 & PostEnc MC \\
    PM-Dom & 93.47 & 93.20 & PostDec MC \\
    PM-Lang & 95.52 & 95.67 & PostDec AED \\
    \bottomrule
  \end{tabular*}
\end{table}

\subsubsection{Language of pretraining data}
\vspace{-3pt}

In order to examine the importance of the multilingual Adaptor pretraining,
we experiment with the PostEnc MC approach but replace
the training data with 1000 hours of English recordings
sampled from the Common Voice corpus. The results are displayed
in Table \ref{tab:results_data_language}. While the differences
with the multilingual pretraining are rather small,
the results confirm that the Adaptor pretraining should at least
include the languages corresponding to the downstream SLU tasks.
Moreover, the spoken \gls{ner} model on the English recordings is also
improved by the multilingual Adaptor pretraining, what can be attributed
to the relatively large amount of the loanwords in this task.

\begin{table}[]
    \caption{Effect of the languages of Adaptor pretraining data (PostEnc \gls{mc} loss).}
    \label{tab:results_data_language}
    \center
    \footnotesize
\vspace{-6pt}
    \begin{tabular*}{0.68\columnwidth}{l | l | c  c }
    \toprule
    \multirow{2}{*}{Dataset} & Metrics  & \multicolumn{2}{c}{Pretraining languages}  \\
    \cmidrule{3-4}
      &    & English & Multiple  \\
    \midrule
    \multirow{2}{*}{SLURP} & IC Acc.$\uparrow$ & \textbf{88.62} & 88.13 \\
                           & SLU-F1$\uparrow$ & \textbf{79.09} & 78.91 \\
    \midrule
    \multirow{2}{*}{SLUE} & F1$\uparrow$ &   \textbf{72.06} & 71.93 \\
                          & label-F1$\uparrow$ & 84.40 & \textbf{85.10} \\
    \midrule
    \multirow{2}{*}{CATSLU} & Acc.$\uparrow$ & 47.08  & \textbf{47.50}  \\
                            & F1$\uparrow$ & 62.43 & \textbf{62.65} \\
    \midrule
    \multirow{2}{*}{MEDIA} & CER$\downarrow$ & 12.88 &  \textbf{12.31}  \\
                           & CVER$\downarrow$ & 15.75 & \textbf{15.07} \\
    \midrule
    \multirow{2}{*}{PM-Dom} & CER$\downarrow$ & \textbf{19.65} & 19.67  \\
                            & CVER$\downarrow$ & 23.55 & \textbf{22.81} \\
    \midrule
    \multirow{2}{*}{PM-Lang} & CER$\downarrow$ &  23.71  &  \textbf{23.36} \\
                             & CVER$\downarrow$ & 27.83  & \textbf{27.22} \\
    \bottomrule
  \end{tabular*}
\vspace{-12pt}%
\end{table}

\vspace{-12pt}
\subsubsection{Scaling up Adaptor pretraining}
\vspace{-3pt}

We select the best performing configuration (PostDec AED)
and run the Adaptor pretraining on our full ASR dataset
comprising of 7000 hours.
It can be seen from Table \ref{tab:results_data_amount}
that the additional pretraining data improves the results
in almost all cases indicating a scaling potential of our method.
In order to compare our E2E SLU with the pipeline approach,
we transcribe the SLU datasets using the PostDec AED model trained on 7000 hours
and subsequently fine-tune and evaluate the NLU models in these ASR transcriptions.
As the comparison
with the  pipeline results in Table~\ref{tab:results_data_amount} shows,
our E2E SLU approach can offer more accurate predictions given the same
training data and pretrained models as the pipeline.
Moreover, Table \ref{tab:results_data_amount} shows the comparison
of our best result with the previous work.
The proposed approach demonstrates overall competitive results with the exception of CATSLU dataset.
Our systems outperform the previous results on four datasets
according to the metrics that evaluate predictions with variable length and large vocabulary (SLU-F1, CVER),
and only on two datasets if we take into account the metrics that evaluate either a single element
from a small predefined set of values (IC accuracy) or a sequence of such elements (CER).
This suggests that for tasks with a larger prediction space, the generative capabilities of our model are more important.

\begin{table}[]
    \caption{
    		Scaling up the Adaptor PostDec AED pretraining with additional data
    		and comparison with the pipeline approach and with the prior work.
	}
    \label{tab:results_data_amount}
    \center
    \footnotesize
\vspace{-6pt}
    \begin{tabular*}{0.97\columnwidth}{l | l |  c  c | c | c  c }
    \toprule
    Dataset & Metrics  & \multicolumn{2}{c|}{Data, hours} & Pipeline & \multicolumn{2}{c}{Prior work} \\
    \cmidrule{3-4}
            &          & 1K & 7K & &  &  \\
    \midrule
    \multirow{2}{*}{SLURP} & IC Acc.$\uparrow$ & 89.33 & 90.04 & 64.88 & \textbf{90.07} & \multirow{2}{*}{\cite{xu2023efficient}}  \\
                           & SLU-F1$\uparrow$  & 80.08 & \textbf{80.66} & 54.78 & 79.90 & \\
    \midrule
    \multirow{2}{*}{SLUE} & F1$\uparrow$ & 72.94 & 75.47  & 63.57 & \textbf{77.20}  & \multirow{2}{*}{\cite{peng2023study}} \\
                          & label-F1$\uparrow$ & 86.57 & 88.14 & 76.66 & \textbf{88.70}  & \\
    \midrule
    \multirow{2}{*}{CATSLU} & Acc.$\uparrow$ & 54.54 & 56.34 & 38.46 & \textbf{86.30} & \multirow{2}{*}{\cite{wang2022arobert}}  \\
                            & F1$\uparrow$ & 69.24 & 71.07  & 58.49 & \textbf{92.56}  &  \\
    \midrule
    \multirow{2}{*}{MEDIA} & CER$\downarrow$ & 13.64 & 12.07 & 29.51 & \textbf{11.20} & \multirow{2}{*}{\cite{ghannay2021we}}  \\
                           & CVER$\downarrow$ & 16.18 & \textbf{14.57}  & 33.45 & 17.20 &   \\
    \midrule
    \multirow{2}{*}{PM-Dom} & CER$\downarrow$ & 18.00  & \textbf{17.90} & 44.80 & 21.90 & \multirow{2}{*}{\cite{caubriere2019curriculum}} \\
                           & CVER$\downarrow$ & 21.27 &  \textbf{21.08} & 51.52 & 35.90 &  \\
    \midrule
    \multirow{2}{*}{PM-Lang} & CER$\downarrow$ & 21.88  & \textbf{21.50} & 49.53 & 26.18 &  \multirow{2}{*}{\cite{caubriere2019curriculum}} \\
                             & CVER$\downarrow$ & \textbf{25.04}  & \textbf{25.04} & 54.62 & 39.28 &   \\
    \bottomrule
  \end{tabular*}
\end{table}

\vspace{-9pt}
\subsection{Cross-lingual SLU}
\vspace{-3pt}

Our SLU approach is mostly multilingual, so the model is highly likely to have
cross-lingual capabilities.
The PortMEDIA-Language benchmark is specifically designed
to evaluate this aspect because it uses the same \gls{sf} tags as
the MEDIA corpus, but is recorded in Italian instead of French.
Similarly to \cite{laperriere2023use}, we experiment with our best
MEDIA model and evaluate it on PortMEDIA-Language test set without any modification
and after fine-tuning on the PortMEDIA-Language training set.
The results shown in Table \ref{tab:results_transfer} outperform
by a large margin most of the zero-shot results reported in Table 3 of \cite{laperriere2023use} and
all of the fine-tuning results reported in Table 5 of \cite{laperriere2023use}.
On top of that, our fine-tuning numbers are better than
any previously reported.

\begin{table}[]
    \vspace{-12pt}
    \caption{Results of transfer learning from MEDIA (French) to PM-Lang (Italian).}
    \label{tab:results_transfer}
    \center
    \footnotesize
\vspace{-6pt}
    \begin{tabular*}{0.76\columnwidth}{l | c  c  c }
    \toprule
    Metrics & No transfer & Zero-shot & Fine-tuning \\
    \midrule
    CER$\downarrow$   & 21.50  & 64.93 & \textbf{20.30} \\
    CVER$\downarrow$  & 25.04  & 71.64 & \textbf{23.65} \\
    \bottomrule
  \end{tabular*}
\vspace{-12pt}%
\end{table}
 
\vspace{-6pt}
\section{Conclusions}
\vspace{-6pt}

We propose a unified \gls{e2e} \gls{slu} approach based on the multilingual \gls{ssl}
pretrained speech and text-to-text models and evaluate it on multiple \gls{slu}
benchmarks. The evaluation results are comparable to or better than the previously reported,
but apply to a more diverse set of the SLU benchmarks.
The pretraining on the medium amount of \gls{asr} data using
the popular \gls{ctc} and \gls{aed} and the novel \gls{mc} approaches helps to improve
the scores across the board and allows outperforming the best-known configurations
in several cases, suggesting that the proposed model can be improved further.


\section{References}
{
\printbibliography
}

\end{document}

%% file: bibliography.tex
\usepackage[
backend=biber,
style=ieee,
citestyle=numeric-comp,
maxbibnames=20,
maxcitenames=20,
doi=false,isbn=false,url=false,eprint=false
]{biblatex} 

\defbibheading{bibliography}[\refname]{}

\DeclareSourcemap{
	\maps[datatype=bibtex, overwrite=true]{
		\map{
			\step[fieldsource=booktitle,
			match=\regexp{.*Interspeech.*},
			replace={Proc. Interspeech}]
			\step[fieldsource=journal,
			match=\regexp{.*INTERSPEECH.*},
			replace={Proc. Interspeech}]
			\step[fieldsource=booktitle,
			match=\regexp{.*ICASSP.*},
			replace={Proc. ICASSP}]
			\step[fieldsource=booktitle,
			match=\regexp{.*icassp_inpress.*},
			replace={Proc. ICASSP (in press)}]
			\step[fieldsource=booktitle,
			match=\regexp{.*Acoustics,.*Speech.*and.*Signal.*Processing.*},
			replace={Proc. ICASSP}]
			\step[fieldsource=booktitle,
			match=\regexp{.*International.*Conference.*on.*Learning.*Representations.*},
			replace={Proc. ICLR}]
			\step[fieldsource=booktitle,
			match=\regexp{.*International.*Conference.*on.*Computational.*Linguistics.*},
			replace={Proc. COLING}]
			\step[fieldsource=booktitle,
			match=\regexp{.*SIGdial.*Meeting.*on.*Discourse.*and.*Dialogue.*},
			replace={Proc. SIGDIAL}]
			\step[fieldsource=booktitle,
			match=\regexp{.*International.*Conference.*on.*Machine.*Learning.*},
			replace={Proc. ICML}]
			\step[fieldsource=booktitle,
			match=\regexp{.*North.*American.*Chapter.*of.*the.*Association.*for.*Computational.*Linguistics:.*Human.*Language.*Technologies.*},
			replace={Proc. NAACL}]
			\step[fieldsource=booktitle,
			match=\regexp{.*Empirical.*Methods.*in.*Natural.*Language.*Processing.*},
			replace={Proc. EMNLP}]
			\step[fieldsource=booktitle,
			match=\regexp{.*Association.*for.*Computational.*Linguistics.*},
			replace={Proc. ACL}]
			\step[fieldsource=booktitle,
			match=\regexp{.*Automatic.*Speech.*Recognition.*and.*Understanding.*},
			replace={Proc. ASRU}]
			\step[fieldsource=booktitle,
			match=\regexp{.*Spoken.*Language.*Technology.*},
			replace={Proc. SLT}]
			\step[fieldsource=booktitle,
			match=\regexp{.*Speech.*Synthesis.*Workshop.*},
			replace={Proc. SSW}]
			\step[fieldsource=booktitle,
			match=\regexp{.*workshop.*on.*speech.*synthesis.*},
			replace={Proc. SSW}]
			\step[fieldsource=booktitle,
			match=\regexp{.*Advances.*in.*neural.*information.*processing.*},
			replace={Proc. NeurIPS}]
			\step[fieldsource=booktitle,
			match=\regexp{.*Advances.*in.*Neural.*Information.*Processing.*},
			replace={Proc. NeurIPS}]
			\step[fieldsource=booktitle,
			match=\regexp{.*Workshop.*on.* Applications.* of.* Signal.*Processing.*to.*Audio.*and.*Acoustics.*},
			replace={Proc. WASPAA}]
			\step[fieldsource=publisher,
			match=\regexp{.+},
			replace={{}}]
			\step[fieldsource=month,
			match=\regexp{.+},
			replace={{}}]
			\step[fieldsource=location,
			match=\regexp{.+},
			replace={{}}]
			\step[fieldsource=address,
			match=\regexp{.+},
			replace={{}}]
			\step[fieldsource=organization,
			match=\regexp{.+},
			replace={{}}]
		}
	}
}